# Kernel Transform Learning

Jyoti Maggu[a] and Angshul Majumdar[a] ∗

[a]*Indraprastha Institute of Information Technology, Okhla Phase 3, New Delhi, 110020, India*

ABSTRACT

This work proposes kernel transform learning. The idea of dictionary learning is well known; it is a synthesis formulation where a basis is learnt along with the coefficients so as to generate / synthesize the data. Transform learning is its analysis equivalent; the transforms operates / analyses on the data to generate the coefficients. The concept of kernel dictionary learning has been introduced in the recent past, where the dictionary is represented as a linear combination of non-linear version of the data. Its success has been showcased in feature extraction. In this work we propose to kernelize transform learning on line similar to kernel dictionary learning. An efficient solution for kernel transform learning has been proposed – especially for problems where the number of samples is much larger than the dimensionality of the input samples making the kernel matrix very high dimensional. Kernel transform learning has been compared with other representation learning tools like autoencoder, restricted Boltzmann machine as well as with dictionary learning (and its kernelized version). Our proposed kernel transform learning yields better results than all the aforesaid techniques; experiments have been carried out on benchmark databases.

---

∗ Corresponding author. Tel.: +91-11-2690-7451; fax: +91-11-2690-7405; e-mail: angshul@iiitd.ac.in

# 1. Introduction

Dictionary learning has garnered immense popularity in image processing and computer vision in the last decade. The technique has been known since the late 90s (Olshausen and Field, 1997; Lee and Seung, 1999); however it was the paper on KSVD by Aharon et al, 2006, that triggered today's popularity on the topic.

Dictionary learning is a synthesis approach, given the data, it learns a dictionary so that it can synthesize / regenerate the data from the learnt coefficients. Researchers in signal processing have used it solve inverse problems. In computer vision, the learnt coefficients are used as features for classification and clustering.

There have been certain extensions to the basic dictionary learning approach. There is the concept of 'double sparsity' (Rubenstein et al, 2010; Lu et al, 2013). It assumes that the dictionary is composed of a linear combination of basis elements from a fixed transform (wavelet, DCT etc.) and learns the linear combination weights. The 'double' arises from the fact that the there is one set of sparse coefficients to define the dictionary and another to define the representing coefficients for the data. Double sparsity does not have any applications in computer vision but has found some applications in inverse problems.

Another extension to dictionary learning is its kernel version. Here it is assumed that in order to represent a non-linear version of the data, the dictionary is formed by a linear combination of the non-linear version of itself. Kernel dictionary learning (Nguyen et al, 2012; Golt and Elad, 2016) does not have any application in signal processing but is useful for vision tasks (Shrivastava et al, 2015).

Transform learning is a recent topic; it is the analysis equivalent of dictionary learning. It learns a transform so that it operates (analyses) on the data to generate coefficients. There are only a handful of papers on this subject of theoretical nature (Ravishankar and Bresler, 2013; Ravishankar et al, 2015; Ravishankar and Bresler, 2015). Ravishankar and Bresler, 2015, showed that it can be applied for the solution of inverse problems such as medical image reconstruction. Shekhar et al, 2014 independently arrived at the same formulation, but dubbed it as 'analysis sparse coding'. There it was shown how coefficients from transform learning can be used for feature extraction.

The concept of 'double sparsity' arising in dictionary learning has been applied to transform learning as well (Ravishankar and Bresler, 2013). Instead of learning a dense transform, it was assumed that the transform is a linear combination of basis elements from a fixed basis (curvelet, DCT, Gabor etc.). Therefore during double sparse transform learning one needed to estimate the linear combination weights defining the transform (from the fixed basis) and the (sparse) transform coefficients.

In this work we introduce kernel transform learning. We apply this technique for unsupervised representation learning. Applications on benchmark classification problems show that our proposed method performs better than other other existing techniques such as dictionary learning and its kernelized version. autoencoder and restricted Botlzmann machine.

# 2. Literature Review

## 2.1. Dictionary Learning

In dictionary learning, we learn a basis and learn the corresponding coefficients from the training data such that the basis / dictionary can synthesize / generate the training data. It was introduced in late 90's as an empirical tool to learn filters (Olshausen and Field, 1997; Lee and Seung, 1999). The usual understanding of dictionary learning is shown in Fig. 1.

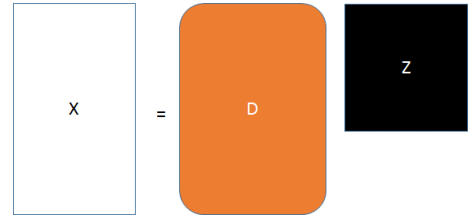

**Fig. 1.** Schematic Diagram for Dictionary Learning

The dictionary ($D$) and the coefficients ($Z$) are learnt from the data ($X$) such that the learnt dictionary and the coefficients can synthesize the data. Mathematically this is represented as,

$$X = DZ \qquad (1)$$

Early studies in dictionary learning focused on learning a basis for representation. There were no constraints on the dictionary atoms or on the loading coefficients. The method of optimal directions (Engan et al, 1999) was used to learn the basis:

$$\min_{D,Z} \|X - DZ\|_F^2 \qquad (2)$$

Here, $X$ is the training data, $D$ is the dictionary to be learnt and $Z$ consists of the loading coefficients. This (2) is solved using alternating least squares, i.e. The coefficients are updated assuming the dictionary is fixed and vice versa.

$$Z_k \leftarrow \min_Z \|X - D_{k-1}Z\|_F^2 \qquad (3a)$$

$$D_k \leftarrow \min_D \|X - DZ_k\|_F^2 \qquad (3b)$$

For problems in sparse representation, the objective is to learn a basis that can represent the samples in a sparse fashion, i.e. $Z$ needs to be sparse. K-SVD (Aharon et al, 2006) is the most well-known technique for solving this problem. Fundamentally, it solves a problem of the form:

$$\min_{D,Z} \|X - DZ\|_F^2 \text{ such that } \|Z\|_0 \leq \tau \qquad (4)$$

Here we have abused the notation slightly, the $l_0$-norm is defined on the vectorized version of Z. The problem with K-SVD is that it is slow, since it requires computing the SVD in every iteration and updating the coefficients via orthogonal matching pursuit.

Dictionary learning finds applications in inverse problems like denoising (Zhang et al, 2013; Kuang et al, 2014) and reconstruction (Ravishankar and Bresler, 2011). It also finds a variety of applications in computer vision where the learnt coefficients are used as features (Mairal, 2010). In recent times, the technique is finding applications in more sophisticated problems of domain adaptation (Fernando et al, 2015; Mudunuri and Biswas, 2016).

Usually dictionary learning solves for a dense dictionary. Such a matrix is difficult to store and operate with. It cannot be applied on large scale data (without extracting patches). To address this issue the concept of double sparsity was introduced (Rubenstein et al, 2010). Here the dictionary is not fully learnt; it is assumed that the dictionary can be expressed as a sparse linear combination of a fixed basis $\Phi$ (e.g. wavelet, DCT etc.), i.e.

$$D = \Phi A \qquad (5)$$

Here $A$ are the sparse weights combining the atoms of the fixed basis to generate the dictionary. The full formulation is therefore (combining with (1)),

$$X = \Phi AZ \qquad (6)$$

The learning is formulated as,

$$\min_{A,Z} \|X - \Phi AZ\|_F^2 \text{ such that } \|Z\|_0 \leq \tau, \ \|A\|_0 \leq \gamma \qquad (7)$$

The idea of decomposing the dictionary into a fixed portion and a learned variable stems from the concept of double sparsity. In kernel dictionary learning this is extended.

*2.2. Kernel Dictionary Learning*

Kernel dictionary learning (Shrivastava et al, 2015; Golts and Elad, 2016) extends the concept of double sparsity in the sense that it defines a dictionary in terms of a linear combination of non-linear version of the data. The formulation is given by,

$$\varphi(X) = \varphi(X)AZ \qquad (8)$$

The non-linear transformation on the data $\varphi(X)$ is synthesized from the coefficients ($Z$) from a dictionary formed by linear combination of its elements – $\varphi(X)A$. The formulation for learning is,

$$\min_{A,Z} \|\varphi(X) - \varphi(X)AZ\|_F^2 \text{ such that } \|Z\|_0 \leq \tau \qquad (9)$$

The problem is solved using alternate minimization. The update step for A is actually independent of the data, since

$$A \leftarrow \min_A \|\varphi(X)(I - AZ)\|_F^2$$
$$A = Z^T(Z^TZ)^{-1} \qquad (10)$$

The update for the sparse coding stage is a solved problem via Kernel Matching Pursuit (Vincent and Bengio, 2002). It basically kernelizes the step of computing the correlation between the non-linear basis and the non-linear residual. However the straightforward Kernel Matching Pursuit requires storage and manipulation of the high dimensional kernel matrix (dimensionality of the number of samples); this is expensive. In order to ameliorate the computational challenges the Nystrom method is used by Golts and Eldar, 2016.

*2.3. Transform Learning*

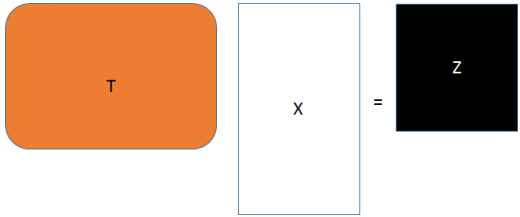

**Fig. 1.** Schematic Diagram for Transform Learning

As mentioned before, transform learning analyses the data by learning a transform / basis to produce coefficients. Mathematically this is expressed as,

$$TX = Z \qquad (11)$$

Here $T$ is the transform, $X$ is the data and $Z$ the corresponding coefficients. Relating transform learning to the dictionary learning formulation (1), we see that dictionary learning is an inverse problem while transform learning is a forward problem.

One may be enticed to solve the transform learning problem by formulating,

$$\min_{T,Z} \|TX - Z\|_F^2 + \mu\|Z\|_0 \qquad (12)$$

Unfortunately such a formulation would lead to degenerate solutions; it is easy to verify the trivial solution $T=0$ and $Z=0$. In order to ameliorate this, the following formulation was proposed by Ravishankar and Bresler, 2013 –

$$\min_{T,Z} \|TX - Z\|_F^2 + \lambda\left(\|T\|_F^2 - \log\det T\right) + \mu\|Z\|_0 \qquad (13)$$

The factor $-\log\det T$ imposes a full rank on the learned transform; this prevents the degenerate solution. The additional penalty $\|T\|_F^2$ is to balance scale; without this $-\log\det T$ can keep on increasing producing degenerate results in the other extreme.

Just as in dictionary learning, one needs to solve (13) by alternating minimization.

$$Z \leftarrow \min_Z \|TX - Z\|_F^2 + \mu\|Z\|_0 \qquad (14a)$$

$$T \leftarrow \min_T \|TX - Z\|_F^2 + \lambda\left(\|T\|_F^2 - \log\det T\right) \qquad (14b)$$

Updating the coefficients (14a) is straightforward. It can be updated via one step of Hard Thresholding [44]. This is expressed as,

$$Z \leftarrow (abs(TX) \geq \mu) \odot \qquad (15)$$

Here $\odot$ indicates element-wise product.

For updating the transform, one can notice that the gradients for different terms in (14b) are easy to compute. Ignoring the constants this is given by –

$$\nabla\|TX - Z\|_F^2 = X^T(TX - Z)$$

$$\nabla\|T\|_F^2 = T$$

$$\nabla\log\det T = T^{-T}$$

In the initial paper on transform learning, a non-linear conjugate gradient based technique was proposed to solve the transform update (Ravishankar and Bresler, 2013. In the second paper (Ravishankar et al, 2015), with some linear algebraic tricks they were able to show that a closed form update exists for the transform.

$$XX^T + \lambda I = LL^T \qquad (16a)$$

$$L^{-1}YX^T = USV^T \qquad (16b)$$

$$T = 0.5R\left(S + (S^2 + 2\lambda I)^{1/2}\right)Q^T L^{-1} \qquad (16c)$$

The first step is to compute the Cholesky decomposition; the decomposition exists since $XX^T + \lambda I$ is symmetric positive definite. The next step is to compute the full SVD. The final step is the update step. One must notice that $L^{-1}$ is easy to compute since it is a lower triangular matrix.

The proof for convergence of such an update algorithm can be found in Ravishankar and Bresler, 2015. It was found that the transform learning was robust to initialization conditions.

Transform learning is new – started in 2015. Dictionary learning is almost a decade older than transform learning. The full potential of transform learning is yet to be understood. There are only a handful papers on this topic. There is only a single paper on its application in reconstruction (Ravishankar and Bresler, 2015) and one paper in its application as an automated feature extraction tool (Shekhar et al, 2014).

**3. Kernel Transform Learning**

In kernel dictionary learning the dictionary is defined as a linear combination of a non-linear representation of the data. This is represented as,

$$\varphi(X) = \underbrace{\qquad} \qquad (17)$$

Here $\varphi$ is for the non-linearity and $Z$ are the coefficients.

This is a synthesis formulation, where the kernel dictionary synthesizes the non-linear representation of the data along with coefficients $Z$. Transform learning is an analysis formulation. Our proposed formulation for the kernelized version of it is expressed as,

$$\underbrace{\qquad}_{transform} = Z \qquad (18)$$

One can notice the similarity and contrast between (18) and (8). In (8), kernel could only be defined while solving the inverse problem. The advantage of our proposed formulation is that, one can define the kernel upfront,

$$K(X,X) = \varphi(X)^T \varphi(X) \qquad (19)$$

This allows expressing (18) as,

$$BK(X,X) = Z \qquad (20)$$

Comparison between transform learning (11) and our proposed formulation (20) is that instead of the data matrix, we have the kernelized data matrix. The usual constraints of transform learning will apply. We formulate the learning as,

$$\min_{B,Z} \|BK(X,X) - Z\|_F^2 + \lambda\left(\|B\|_F^2 - \log\det B\right) + \mu\|Z\|_0 \qquad (21)$$

The update for the coefficients remains the same as before (15) –

$$Z \leftarrow (abs(BK(X,X)) \geq \mu) \odot \qquad X) \qquad (22)$$

The update for $B$ remains exactly the same as the update for the transform in (16).

This concludes the training phase. During testing, we need to generate the features for the sample $x_{test}$. In transform learning, this is expressed as: $z_{test} \leftarrow (abs(Tx_{test}) \geq \mu) \odot \qquad$. For Kernel transform learning, the corresponding expression will be:

$$B\varphi(X)^T \varphi(x_{test}) = z_{test} \qquad (23)$$

The Kernel is automatically defined –

$$K(X,x_{test}) = \varphi(X)^T \varphi(x_{test}) \qquad (24)$$

This allows expressing (18) as follows,

$$BK(X,x_{test}) = z_{test} \qquad (25)$$

Since, we look for sparse features, we need to solve,

$$\min_{z_{test}} \|BK(X,x_{test}) - z_{test}\|_2^2 + \mu\|z_{test}\|_0 \qquad (26)$$

The closed form update is

$$z_{test} \leftarrow (abs(BK(X,x_{test})) \geq \mu) \odot \qquad x_{test}) \qquad (27)$$

This concludes the test feature extraction. One can immediately notice that transform learning / kernel transform learning will be significantly faster than dictionary learning and its kernelized version during operation testing. This is because, dictionary learning requires solving an iterative optimization problem (OMP or iterative soft thresholding) as opposed to a simple closed form solution (hard thresholding) in transform learning.

*3.1. Efficient Implementation*

The Kernel matrix $K(X,X)$ is positive semidefinite. Let us assume that the dimensionality of the data is n x 1 and there are $N$ training samples. Then the dimensionality of the kernel matrix is N x N. If the data is high dimensional, i.e. $n > N$, the Kernel matrix is smaller than the data matrix (n x N). Therefore solving the Kernelized version directly is computationally sensible. But when the number of samples is much larger than the dimensionality of the data, storing and manipulating the Kernel matrix is challenging in terms of memory. Therefore we need a more efficient implementation.

The kernel matrix can be factored by eigen decomposition.

$$K(X,X) = U_{N \times n} \Lambda_{n \times n} U_{N \times n}^T \qquad (28)$$

where $U$ is the eigenvector matrix and $\Lambda$ the diagonal matrix of eigenvalues sorted in descending order. Computing the eigen decomposition is expensive, but we will need to do it only once.

With the eigen decomposition, (28) can be expressed as follows,

$$BU\Lambda = UZ \qquad (29)$$

The exact solution to this problem would be,

$$\min_{B,Z} \|BU\Lambda - Z'\|_F^2 + \lambda\left(\|B\|_F^2 - \log\det B\right) + \mu\|UZ\|_0 \qquad (30)$$

Alternating minimization of (30) leads to,

$$B \leftarrow \min_B \|BU\Lambda - Z'\|_F^2 + \lambda\left(\|B\|_F^2 - \log\det B\right) \qquad (31a)$$

$$Z' \leftarrow \min_{Z'} \|BU\Lambda - Z'\|_F^2 + \mu\|UZ\|_0 \qquad (31b)$$

Solving (31a) is straightforward; it is the same as the standard transform update; here $U\Lambda$ plays the role of data.

Solving (31b) is not the same as the update for the coefficients in transform learning. In this work, we show how (31b) can be efficiently solved using variable splitting.

In this work we introduce a proxy variable $P=UZ$. Relaxing the equality constraint of the ensuing augmented Lagrangian by the Bregman variable leads to:

$$\min_{Z',P} \|BU\Lambda - Z'\|_F^2 + \mu\|P\|_0 + \varepsilon\|P - UZ - B\|_F^2 \qquad (32)$$

By using the alternating directions method of multipliers (32) can be broken down into the following simpler sub-problems.

$$P1: \min_{Z'} \|BU\Lambda - Z'\|_F^2 + \varepsilon\|P - UZ' - B\|_F^2$$

$$P2: \min_P \mu\|P\|_0 + \varepsilon\|P - UZ - B\|_F^2$$

Subproblem P1 is a simple least squares problem having a closed form solution in the form of pseudoinverse. Subproblem P2 has a closed form solution in the form of hard thresholding.

This concludes the training. The formulation is memory efficient; instead of working with the high dimensional kernel matrix, we can work with its eigen decomposition which makes the problem of size similar to linear transform learning. Thus the memory requirement is significantly reduced.

However there is a trade-off, the time required for solving this will be larger. This is because instead of the closed form update (22), we need to solve an iterative problem (31b). This is the price we pay (in terms of increased training time) to gain in memory efficiency.

During testing, computing the eigen decomposition is not required, it only increases the computational requirement. One can simply compute the kernel matrix $K(X, x_{test})$ and with the value of $B$ obtained during training, the feature for the test sample can be computed as before (27). Thus, even though our implementation increases the training time, the test time remains unaffected. It is the same as that of linear dictionary learning.

## 4. Experimental Results

### 4.1. Datasets

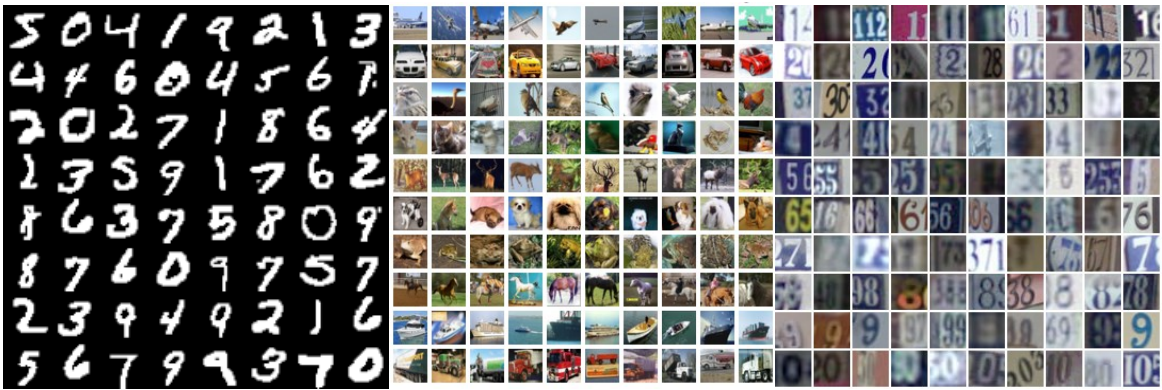

**Fig. 3.** Left to Right: Samples from MNIST, CIFAR-10 and SVHN

**Table 1. Comparison of Accuracy on Benchmark Datasets**

| Classifier | Dataset | AE | RBM | DL | KDL | TL | KTL | eKTL |
|---|---|---|---|---|---|---|---|---|
| NN | MNIST | 95.31 | 94.55 | 93.39 | 94.51 | 94.70 | **95.82** | **95.80** |
|  | CIFAR-10 | 79.45 | 76.83 | 80.27 | 81.19 | 81.61 | **82.09** | **82.09** |
|  | SVHN | 85.79 | 85.08 | 92.11 | 92.47 | 92.42 | - | **92.99** |
| SVM | MNIST | 96.62 | 96.50 | 97.56 | 98.22 | 98.08 | **98.76** | **98.74** |
|  | CIFAR-10 | 80.19 | 78.55 | 82.54 | 83.18 | 83.39 | **83.97** | **83.97** |
|  | SVHN | 87.52 | 87.88 | 93.50 | 93.79 | 93.61 | - | **94.05** |
| ANN | MNIST | 94.20 | 91.80 | 96.49 | 97.08 | 97.31 | **97.99** | **97.99** |
|  | CIFAR-10 | 82.56 | 82.95 | 82.07 | 82.69 | 82.84 | **83.19** | **83.17** |
|  | SVHN | 92.77 | 93.45 | 93.07 | 93.86 | 93.64 | - | **94.20** |

**Table 2. Comparison of Training Time in Seconds**

| Dataset | AE | RBM | DL | KDL | TL | KTL | eKTL |
|---|---|---|---|---|---|---|---|
| MNIST | 290 | 470 | 1100 | 1205 | 259 | 967 | 270 |
| CIFAR-10 | 316 | 514 | 1209 | 1398 | 278 | 1102 | 301 |
| SVHN | 21068 | 6409 | 7568 | 8010 | 2021 | - | 2198 |

In this work we have tested on three benchmark representation learning datasets. The first one is the MNIST. The MNIST digit classification task is composed of 28x28 images of the 10 handwritten digits. There are 60,000 training images with 10,000 test images in this benchmark.

The CIFAR-10 dataset is composed of 10 classes of natural images with 50,000 training examples in total, 5,000 per class. Each image is an RGB image of size 32x32. These images need to be preprocessed. We follow the standard preprocessing technique – the RGB is converted to YUV and the Y channel is used. Before putting it for training / testing, mean subtraction and global contrast normalization is done.

The Street View House Numbers (SVHN) dataset is composed of 604,388 images (using both the difficult training set and simpler extra set) and 26,032 test images. The goal of this task is to classify the digit in the center of each cropped 32x32 color image. We preprocessed these samples in the same way as CIFAR.

The first two are moderately large datasets; but the kernel matrices can be operated with on a personal computer. However the kernel matrix corresponding to the last dataset cannot be directly operated upon; one needs the efficient implementation.

### 4.2. Classification Accuracy

In this work we compare the feature extraction performance of our proposed kernel transform learning (KTL) with transform learning (TL). There are two version of kernel transform learning; the first one is the direct one and will be dubbed as KTL; the second one is the efficient implementation and will be dubbed as eKTL. Since transform learning is the analysis version of dictionary learning, we compare dictionary learning (DL) and kernel dictionary learning (KDL) as well. The efficient version of the kernel dictionary learning from Golts and Elad, 2016 is used. Comparison is also performed with other popular representation learning tools – autoencoder (AE) and restricted Boltzmann machine (RBM).

The tools we compare against are unsupervised representation learning methods. For classification we use three popular techniques – nearest neighbour (NN), support vector machine (SVM) and artificial neural network (ANN). For DL and KDL, 700 atoms yields the best results (Golts and Elad, 2016). For TL and its kernel versions 200 atoms were found to yield the best overall results. For both kernel dictionary learning and kernel transform learning a polynomial kernel of order 4 has been used. For AE and RBM, the number of nodes are 240; this yields the best results for all the datasets. The results are shown in Table 1.

We find that our proposed kernel transform learning yields the best results without fail. The memory efficient implementation is sometimes marginally worse than direct version, but yields results better than all other representation learning tools (compared against) nevertheless. For the very large SVHN training dataset, it is not possible to carry out the straightforward kernel implementation; therefore we have not showed the results.

*4.3. Time Comparison*

We compare the training time for the different methods. These are the times required for training the different representation learning tools and not the classifiers. Experiments have been carried out on a computer running on am Apple MacBook Air Core i5 5th Gen with 8 GB RAM, 128 GB SSD having Mac OS Sierra. The results are shown in Table 1.

We find that transform is in general faster than dictionary learning. In fact it is the fastest of all representation learning techniques. It must be remembered that more efficient implementations of dictionary learning are available, but we have used the gold standard KSVD algorithm (Aharon et al, 2006). The kernelized version of dictionary learning takes slightly more time than the base algorithm, this is because it requires computing eigen decompositions.

Our proposed basic kernel transform learning method is significantly slower than the original transform learning; this is because the Kernel matrix is much larger than the data matrix. But with our proposed efficient implementation, we reduce the training time significantly; it is only slightly slower than the linear transform learning formulation.

## 5. Conclusion

This work proposes a kernelized version of transform learning. The direct version is memory intensive for problems where the number of training samples far exceed the dimensionality of the data; for such problems we have proposed an efficient version which has almost the same complexity as that of the linear transform learning.

The proposed technique has been employed as a representation learning tool. The technique was compared with standard representation learning tools like autoencoder and restricted Botlzmann machine as well as with dictionary learning, kernel dictionary learning and transform learning on benchmark datasets. Experimental results showed that our proposed kernel dictionary learning always yields the best results in terms of accuracy. In terms of speed, the efficient version of kernel transform learning yields results which are only slightly worse than the direct version (better than all others), but significantly faster.

## Acknowledgments

This work is partially supported by the Infosys Center for Artificial Intelligence @ IIIT Delhi.

## References

Aharon, M., Elad, M., Bruckstein, A., 2006. K-SVD: An Algorithm for Designing Overcomplete Dictionaries for Sparse Representation. IEEE Transactions on Signal Processing. 4311-4322.

Engan, K., Aase, S., Hakon-Husoy, J., 1999. Method of optimal directions for frame design. IEEE International Conference on Acoustics, Speech, and Signal Processing.

Fernando, B., Tommasi, T., Tuytelaars, T. 2015. Joint cross-domain classification and subspace learning for unsupervised adaptation. Pattern Recognition Letters, 65, 60-66.

Golts, A., and Elad, M. 2016. Linearized Kernel Dictionary Learning. IEEE Journal of Selected Topics in Signal Processing, 10, 4, 726-739.

Kuang, Y., Zhang, L., Yi, Z. 2014. An adaptive rank-sparsity k-svd algorithm for image sequence denoising. Pattern Recognition Letters. 45, 46-54.

Lee, D. D., Seung, H. S. 1999. Learning the parts of objects by non-negative matrix factorization. Nature, 401(6755), pp. 788-791.

Lu, X., Yuan, Y., Yan, P. 2013. Image super-resolution via double sparsity regularized manifold learning. IEEE transactions on circuits and systems for video technology. 23, 2022-2033.

Mairal, J. 2010. Sparse coding for machine learning, image processing and computer vision. Doctoral dissertation, Cachan, Ecole normale supérieure.

Mudunuri, S. P., Biswas, S. 2016. A coupled discriminative dictionary and transformation learning approach with applications to cross domain matching. Pattern Recognition Letters. 71, 38-44.

Nguyen, H. V., Patel, V. M., Nasrabadi, N. M., Chellappa, R. 2012. Kernel dictionary learning. IEEE ICASSP, pp. 2021-2024.

Olshausen, B., Field, D., 1997. Sparse coding with an overcomplete basis set: a strategy employed by V1? Vision Research. 3311-3325.

Ravishankar, S., Bresler, Y. 2011. MR image reconstruction from highly undersampled k-space data by dictionary learning. IEEE transactions on medical imaging, 30, 1028-1041.

Ravishankar, S., Bresler, Y. 2013. Learning Sparsifying Transforms. IEEE Transactions on Signal Processing. 61, 1072-1086.

Ravishankar, S., Bresler, Y. 2013. Learning doubly sparse transforms for images. IEEE Transactions on Image Processing. 22, 4598-4612.

Ravishankar, S., Wen, B., Bresler, Y. 2015. Online Sparsifying Transform Learning - Part I: Algorithms. IEEE Journal of Selected Topics in Signal Processing. 9, 625-636.

Ravishankar, S., Bresler, Y. 2015. Online Sparsifying Transform Learning - Part II: Convergence Analysis. IEEE Journal of Selected Topics in Signal Processing. 9, 637-746.

Ravishankar, S., Bresler, Y. 2015. Efficient Blind Compressed Sensing using Sparsifying Transforms with Convergence Guarantees and Application to MRI. SIAM Journal on Imaging Sciences, 8, 4, 2519-2557.

Rubinstein, R., Zibulevsky, M., and Elad, M. 2010. Double Sparsity: Learning Sparse Dictionaries for Sparse Signal Approximation. IEEE Transactions on Signal Processing, 58, 1553-1564.

Shekhar, S., Patel, V. M. Chellappa, R. 2014. Analysis sparse coding models for image-based classification. IEEE International Conference on Image Processing. 5207-5211.

Shrivastava, A., Patel, V. M., Chellappa, R. 2015. Non-linear dictionary learning with partially labeled data. Pattern Recognition. 48, 3283-3292.

Vincent, P., Bengio, Y. 2002. Kernel Matching Pursuit. Machine Learning. 48, 1, 165–187.

Zhang, X., Feng, X., Wang, W., Liu, G. 2013. Image denoising via 2D dictionary learning and adaptive hard thresholding. Pattern Recognition Let...